\documentclass[letterpaper, 10 pt, conference]{ieeeconf}
\IEEEoverridecommandlockouts
\newcommand{\conv}[1]{$\left[\begin{array}{ll} \text{1}\times \text{1}\times \text{1} \text{ conv}\\ \text{3}\times \text{3}\times \text{3} \text{ conv} \end{array}\right] \times \text{#1}$}

\newcommand{\cross}[1]{#1 $\times$ #1}

\usepackage[utf8]{inputenc}
\usepackage[english]{babel}
\usepackage{floatrow}
\usepackage{graphics} 
\usepackage{epsfig} 
\usepackage{mathptmx} 
\usepackage{times} 
\usepackage{cite}
\usepackage{amsmath} 
\usepackage{amssymb}  
\usepackage{graphicx}
\usepackage{mwe}
\usepackage{lipsum}%
\usepackage{booktabs}
\usepackage{multirow}
\usepackage[algoruled,boxed,lined]{algorithm2e}
\usepackage[mathscr]{euscript}
\usepackage{tabularx}
\usepackage{amsfonts}
\usepackage{url}
\usepackage{bm}
\usepackage{adjustbox}
\usepackage{footnote}
\usepackage[para,online,flushleft]{threeparttable}
\usepackage{multirow}
\title{\LARGE \bf Real-time Intent Prediction of Pedestrians for \\ Autonomous Ground Vehicles via Spatio-Temporal DenseNet}

\author{Khaled Saleh, Mohammed Hossny, and Saeid Nahavandi 
	\thanks{K. Saleh, M. Hossny, and S. Nahavandi are with the 
		Institute for Intelligent Systems Research and Innovation (IISRI), Deakin University, VIC, Australia. E-mails: {\tt \{kaboufar, mhossny, saeid.nahavandi\}@deakin.edu.au}.}
}
\begin{document}
	
	\maketitle
	\thispagestyle{empty}
	\pagestyle{empty}	
	\begin{abstract}

Understanding the behaviors and intentions of humans are one of the main challenges autonomous ground vehicles still faced with.  More specifically, when it comes to complex environments such as urban traffic scenes, inferring the intentions and actions of vulnerable road users such as pedestrians become even harder. In this paper, we address the problem of intent action prediction of pedestrians in urban traffic environments using only image sequences from a monocular RGB camera. We propose a real-time framework that can accurately detect, track and predict the intended actions of pedestrians based on a tracking-by-detection technique in conjunction with a novel spatio-temporal DenseNet model. We trained and evaluated our framework based on real data collected from urban traffic environments. Our framework has shown resilient and competitive results in comparison to other baseline approaches. Overall, we achieved an average precision score of 84.76\% with real-time performance at 20 FPS.
		
\end{abstract}
\section{Introduction}
\begin{figure}[t]
\includegraphics[keepaspectratio=true, width=\columnwidth]{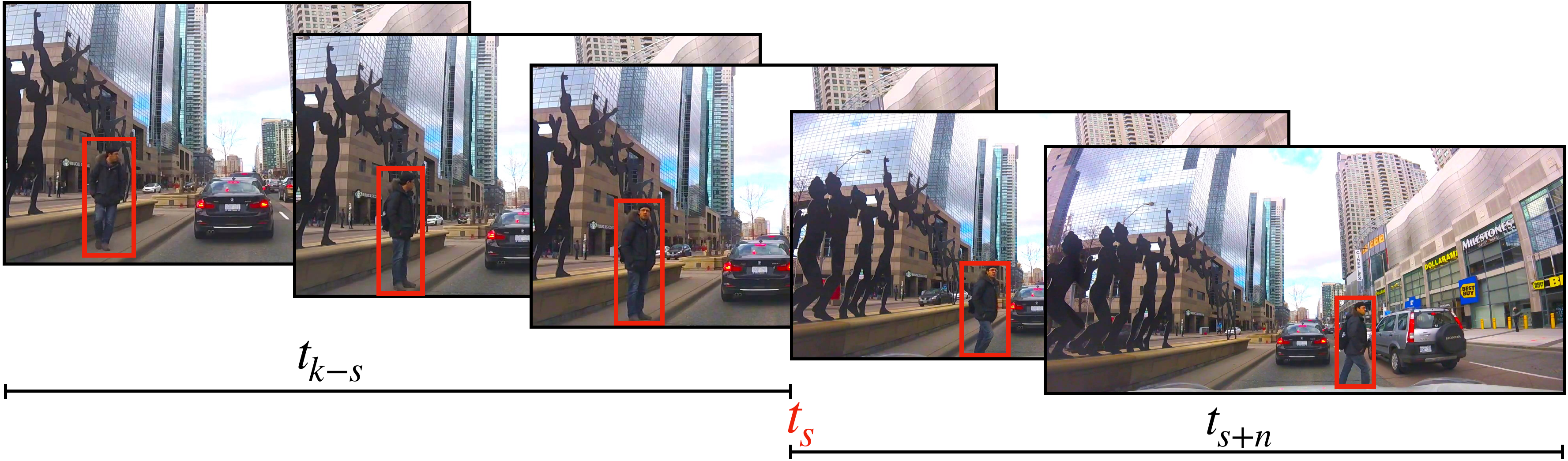}
\caption{Crossing scenario from the newly released JAAD dataset~\cite{rasouli2017they}}
\label{fig:sample}
\end{figure}

The adoption of autonomous ground vehicles (AGVs) in real-life applications have recently got the attention of versatile industries. From food and parcel delivery~\cite{joerss2016parcel} to self-driving vehicles~\cite{Daily2017}, AGVs will be playing a vital role in our day-to-day life in the near future. AGVs are still however encountered with a number of technological hurdles preventing them from the wide adoption and acceptance by the society~\cite{lee2015can}. One of the main challenges that AGVs are still faced with is the lack of deep understanding of the behaviors of humans around them. Humans, on the other hand, can subconsciously understand and predict each others behaviors in various scenarios. For example, the scenario in Fig.~\ref{fig:sample}, any human driver will be expecting that the pedestrian in the scene is probably intending to cross. An autonomous vehicle (AV) on the other hand, might find it really difficult to anticipate this behavior. Thus, the understanding of human behaviors and intentions was considered as one of the most crucial capabilities that AGVs need to acquire~\cite{huang2016anticipatory, saleh2017towards}. 

Our focus in this work will be on the anticipation of human behaviors by AGVs in the context of traffic environments using only one sensor modality (i.e. monocular RGB camera). That being said, the same methodology still can be applied to other environments that involve humans and AGVs such as industrial and indoor environments~\cite{sebanz2009prediction}. In specific, we are in this work will be investigating the behaviors and intentions of one of the vulnerable road users, the pedestrians, that AVs are still challenged with in urban traffic environments~\cite{saleh2017towards}. In the literature, the intent prediction problem from video sequences is often tackled using techniques from the video action/activity recognition domain~\cite{kataoka2016recognition, saleh2017early}. The state-of-the-art frameworks for video action and activity recognition models utilize 2D convolution neural network (ConvNet) in combination with recurrent neural networks (RNNs). As a result, they can model the spatio-temporal information from video streams. Most of these models are however not suitable for real-time applications such as the prediction of intended actions of pedestrians in traffic environments because of the intensive computations required for RNNs. Additionally, these models are not unified in a way that can be easily adapted for a critical task such as intent prediction of pedestrians. For instance, most of these prediction models do not include the pre-processing stages needed for their input data as part of their overall framework. Moreover, they also do not take into account the inherent uncertainty nature of the noisy observations coming from the separate pre-processing stages which commonly exist in the context of AGVs.
Thus, in this work, we are proposing a novel real-time unified framework for pedestrians' intent prediction and localization from video sequences. Our framework utilizes a distinctive architecture of spatio-temporal ConvNet models based on densely connected convolutional networks (DenseNets)~\cite{huang2017densely} to model the spatio-temporal information in video sequences. Spatio-temporal ConvNets have been recently shown to provide a reliable temporal modeling especially for image sequences with comparable or even better results than RNNs~\cite{tran2015learning}. DenseNets, on the other hand, have also shown remarkable results in effectively modeling spatial information in deep 2D ConvNets with less number of parameters to be trained. Our framework also integrates a sub-module for real-time tracking-by-detection of pedestrians from video sequences based on state-of-the-art single ConvNet object detection architecture (YOLOv3~\cite{redmon2018yolov3}) and simple online and real-time tracking algorithm (SORT~\cite{Bewley2016sort}). In summary, the contributions of this work are three folds:
\begin{itemize}
	\item A novel framework for effective prediction of the intended actions pedestrians from video sequences using spatio-temporal DensNet.
	\item An integrated accurate tracking of multi-pedestrians in urban traffic environments as a part of the intent action prediction framework, which takes into account the uncertainty exists in noisy observations and detections from a moving observer. 
	\item A real-time unified and flexible framework for intent prediction of pedestrians in traffic environments using only images sequences coming from an RGB camera. 
\end{itemize} 
	
The next section provides an overview of the related work from the literature. Section~\ref{method}, covers the proposed methodology. Section~\ref{expr}, provides the details of our experiments and the performance evaluation of our proposed framework. Finally, in Section~\ref{concl}, we conclude the paper.

\section{Related Work}\label{related}
The problem of intent and action prediction of pedestrians and humans in general from image sequences has got an increasing interest over the past few years~\cite{rasouli2017they, kataoka2016recognition, saleh2017early}. More specifically, in the context of AGVs in traffic environments, this problem is still an active research area due to its complexity. In this section, we will give an overview of the work from the literature that is related to the intent prediction problem of pedestrians from image sequences specifically in traffic environments. In~\cite{saleh2017early}, a multi-task 2D ConvNet model was introduced for early intent prediction of pedestrians' actions, where the problem was formulated as an image classification problem. Given a single bounding box image of a pedestrian in a traffic scene, they classify whether the pedestrian is walking or standing. Since they were only classifying the actions and not trying to anticipate them, they did not rely on any temporal information such a consecutive sequence of bounding box images. They did not also investigate scenarios of multi pedestrians in the scene. Moreover, they did not consider how the object detection stage could influence the performance of their model as they were relying on ground truth bounding boxes annotation. Similarly, in~\cite{rasouli2017they} they also relied on a 2D ConvNet model based on AlexNet architecture~\cite{krizhevsky2012imagenet} to predict whether pedestrians will cross the road or not. They introduced two different models, one only takes a single bounding box image as in~\cite{saleh2017early}. The other one takes as input a sequence of 15 consecutive bounding boxes of the pedestrians before they cross. Then, they extract the features after the last fully connected layer from AlexNet and use them to train a linear SVM classifier to decide whether the pedestrian will cross or not. They trained their models using images collected from short sequence videos using a vehicle-mounted monocular RGB camera in diverse urban traffic environments. Similar to~\cite{rasouli2017they}, the pedestrian detection phase was not also part of their proposed model and they relied on ground-truth annotations for the bounding boxes of the pedestrians. In~\cite{carreira2017quo}, two of the most common approaches that have been recently heavily utilized in literature for video activity recognition were presented. These two approaches are often referred to in the literature as spatio-temporal approaches. The first approach is based on a 2D ConvNet model which takes $k$-sequence of images and for each image it extract its features after the last fully connected layers similar to~\cite{kataoka2016recognition, rasouli2017they}. However, instead of feeding these features to a classification stage directly, they feed them to a long-short-term memory (LSTM) layer(s) first. LSTM is one of the most commonly used RNN architectures which is efficient in modeling temporal or time-series information in general~\cite{gers1999learning}. 
	\begin{figure*}[t]
		\includegraphics[keepaspectratio=true, width=\textwidth]{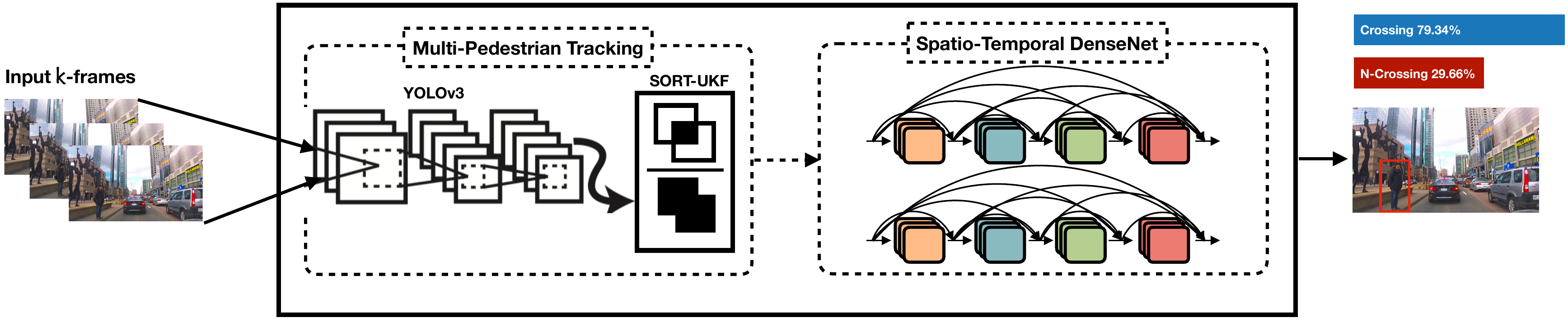}
		\caption{Proposed unified framework for intent action prediction of pedestrians. The input to our framework is a sequence of $k$-frames. We first detect and track all the pedestrians (first stage) in the scene and then predict their intended actions (second stage).}
		\label{fig:frmwrk}
	\end{figure*}	
One drawback of the aforementioned approach is its lack of flexibility because of the two separate stages of training (one for the 2D ConvNet model and the other stage for the LSTM). Thus, another approach based on spatio-temporal ConvNets using 3D convolutions was introduced. Spatio-temporal ConvNets was introduced in~\cite{tran2015learning} as an extension to the conventional 2D ConvNets but with 3D convolution filters for modeling video frames both spatially and temporally. Unlike the aforementioned approach, they can directly in a single stage, model the hierarchical representations of spatio-temporal data from video frames. One of the challenges that the spatio-temporal ConvNet models introduced in~\cite{carreira2017quo, tran2015learning} are faced with is their applicability to only shallow ConvNet architectures. Which in return affects their performance in comparison to 2D ConvNet with LSTMs. In terms of real-time performance, both the aforementioned models (2D ConvNets with LSTM and spatio-temporal ConvNets) require an extensive amount of GPU computations, which in return make them not a very appealing approach to be deployed in AGVs for real-time applications. Thus, till the rest of this paper, we will introduce, discuss and evaluate our proposed approach which combines between the level of accuracy expected from spatio-temporal techniques but with the advantage of low latency at the inference phase. As a result, our proposed approach can be easily adopted for real-time intent action prediction of pedestrians using only sequence of RGB images from a monocular camera.

\section{Proposed Method}~\label{method}
The goal of our work is to provide AGVs with the capabilities to efficiently predict the intended actions of the pedestrians around them in real-time using only video stream from a monocular RGB camera. In fulfilling this goal, we are introducing a unified framework that can firstly detect and track the pedestrians in traffic environments using a moving observer RGB camera. Secondly, our framework takes $k$-sequence of observed bounding boxes of the pedestrians and using a real-time spatio-temporal model it predicts the intended actions of the pedestrians.

\subsection{Problem Formulation}\label{formulation}
In this work, we are focused on specific scenarios of pedestrians in traffic environment which might or not cross the road at any moment. Since in crossing scenarios, the temporal dependency of subsequent frames of pedestrians over time prior to the actual crossing could help in predicting their final intended action. Thus, we found that the spatio-temporal ConvNet approach would be an effective approach for this task. As we have pointed out in Section~\ref{related}, that conventional spatio-temporal ConvNet usually requires a huge amount of parameters for training them which as a result affects their real-time performance at inference time. Therefore, we will be utilizing the DenseNet architecture as the backbone architecture of our spatio-temporal ConvNet model and we refer to it as spatio-temporal DenseNet (ST-DenseNet). The rationale behind choosing DenseNet specifically is because of its efficient number of parameters required for training them as well as their recent state-of-the-art results in image classification tasks. ST-DenseNet has also recently shown promising results in video activity recognition tasks~\cite{diba2018temporal}. The formulation and the setup for our task is however slightly different from video activity recognition tasks. In a video activity recognition task, the assumption is, given a video of a certain human activity $a$ starts at time $t_s$ and ends at time $t_{s+n}$. In light of this scenario according to Fig.~\ref{fig:sample}, the input to the ST-DenseNet for video activity recognition task, would be an image sequence of all the frames of the input video from time $t_s$ to time $t_{s+n}$ inclusive. On the other hand, in the formulation for our intent action prediction task, the input image sequence will be all the $k$ frames before the starting of the activity (i.e from time $t_{k-s}$ to time $t_{s}$). In the following sub-sections, we will discuss the two main stages of our proposed framework (shown in Fig.~\ref{fig:frmwrk}) in more details.

\subsection{Multi-Pedestrian Tracking}\label{multi}
Initially and before we can be able to predict the intended actions of pedestrians in the traffic scene, we need first to detect and track them over a $k$-image sequence. One of the most utilized approaches for this task especially in the context of AGVs is the tracking-by-detection paradigm~\cite{Bewley2016sort,yu2016poi,avidan2004support}. The reason for the wide adoption of the tracking-by-detection paradigm is its balance in giving good accurate estimations without neglecting the real-time performance which is crucial for AGVs. Thus, we will rely on this paradigm for the multi-pedestrian tracking stage of our proposed framework. More specifically, we will utilize a modified version of the simple online and real-time tracking (SORT)~\cite{Bewley2016sort} which recently achieved state-of-the-art results in multiple object tracking (MOT) benchmarks. One of the main advantages of the SORT is its super real-time performance, where it was benchmarked as 20x faster than the other state-of-the-art trackers. SORT exploits the recent success of ConvNet-based models for visual object detection in order to provide frame-to-frame associations. Given bounding boxes (BBoxes) from a generic ConvNet-based object detection model, SORT first predicts the next time step motion of the detected pedestrians' BBoxes using a linear Kalman filter (KF)~\cite{kalman1960new}. Then, it associates the predicted BBoxes with the observed BBoxes from the ConvNet model using a Hungarian method based on the intersection-over-union (IOU) distance~\cite{kuhn1955hungarian}. The following are our two modifications on the original SORT algorithm for our multi-pedestrian tracking stage:
\subsubsection*{\textbf{Detection using YOLOv3}}
Instead of the Faster-RCNN model that was used for the visual object detection in the original SORT algorithm, we will rely on the recently released YOLOv3 architecture for real-time pedestrian tracking~\cite{redmon2018yolov3}. The reason for replacing Faster-RCNN model is because it consists of two-stages for its operation which slows down its run-time. YOLOv3 on the other hand, is one of the state-of-the-art real-time single stage ConvNet-based object detectors. Since the multi-pedestrian tracking stage is a sub task of our framework, hence we need to make sure that it is done as fast as possible. Thus, we found that YOLOv3 given its real-time performance and accurate results would be the perfect match for the requirements of our framework. We utilized the same architecture as the YOLOv3-416 architecture proposed in~\cite{redmon2018yolov3}. Our only modification in the architecture is changing the number of object classes in the final classification stage to our only object of interest (i.e pedestrian). YOLOv3-416 takes as input an RGB image of size $(416H \times 416W)$, hence the name.  
	
\subsubsection*{\textbf{Estimation using UKF}}
In the original SORT algorithm, it relied on the linear KF for estimating the motion model of pedestrians. The underlying assumption for the KF was a linear constant velocity model. Since in traffic environments and more specifically in crossing scenarios, pedestrians do not follow a linear motion model as it was shown in~\cite{keller2011will}. Thus, we are proposing the unscented Kalman filter (UKF)~\cite{wan2000unscented} as a non-linear motion estimation model for the multi-pedestrian tracking stage. Similar to the linear KF of the original SORT, our proposed SORT version using UKF (SORT-UKF) is assuming a constant velocity model for the pedestrians where the state of each pedestrian is represented by the following state $\textbf{x}$:
	
	\begin{equation}
	\textbf{x} = [u, v, s, r, \dot{u}, \dot{v}, \dot{s}]^T
	\end{equation}
	where $u$ as $v$ are the horizontal and vertical location of the center pixel of the pedestrians' BBoxes respectively. While $s$ is the scale area of the BBox and $r$ is the aspect ratio of the BBox.
	We used the same data association technique used in the SORT algorithm, based on the intersection-over-union (IOU) distance. Each detected BBox of pedestrian in the scene is associated based on IOU with the predicted BBoxes from the UKF. Afterwards, the detections are used for updating the UKF.
	\begin{figure}[t]
		\includegraphics[keepaspectratio=true, height= 2.4cm, width=\columnwidth]{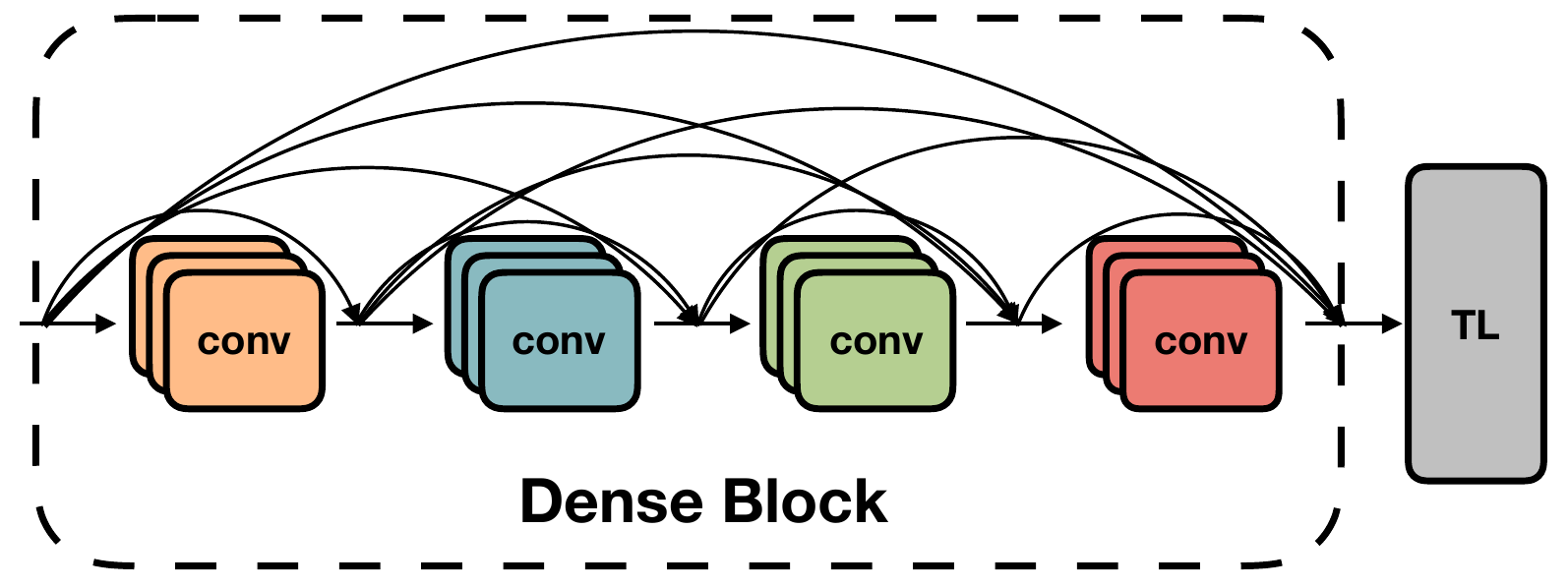}
		\caption{A 4-layer dense block of our ST-DenseNet model. Each ''conv`` is a composition function of (BN-ReLU-3D Convolution). Between each dense block a transition layer (TL) exists.}
		\label{fig:dblock}
	\end{figure}
\subsection{Spatio-Temporal DenseNet (ST-DenseNet)}
In ST-DenseNet, we extend the original DenseNet architecture proposed in~\cite{huang2017densely} by replacing the 2D kernels of the convolution and pooling layers with 3D counterparts. 3D convolution layers convolve its input feature maps spatially similar to the 2D convolution as well as temporally to model the temporal dependency between consecutive frames.  Similarly, 3D pooling layers down-sample the size of its input feature maps spatially and temporally. The kernel of both 3D convolution layers and 3D pooling layers is of size ($s \times s \times d$), where $s$ is the spatial size and $d$ is the input video frames depth/length. The name of the DenseNet is after its architectural design, where each layer in the DenseNet is directly/densely connected to all its preceding layers. This in return helps in improving the flow of information and gradients during the training phase which accelerates it significantly. Additionally, it also reduces the number of parameters needs to be learned because the network can preserve information and eliminates the redundant re-learning of the same weights such as in conventional ConvNet architectures.  The cornerstone unit of DenseNet architecture is the dense blocks. As it can be shown in Fig.~\ref{fig:dblock}, similar to 2D DenseNet, each "conv`` layer in dense block in the ST-DenseNet is internally a composition function of three consecutive operations: batch normalization (BN)~\cite{ioffe2015batch}, ReLU and a 3D convolution (with size $3 \times 3 \times 3$) unlike the 2D one in 2D DenseNet. The following equation describes how the output feature map $\textbf{f}_l$ from layer $l$ in ST-DenseNet's dense block is calculated:
	\begin{equation}
	\textbf{f}_l = \textbf{H}_l([\textbf{f}_0, \textbf{f}_1, \ldots,\textbf{f}_{l-1}])
	\end{equation} 
where $\textbf{H}_l$ is the composition function (BN-ReLU-3DConv). The $[\textbf{f}_0, \textbf{f}_1, \ldots,\textbf{f}_{l-1}]$ is dense connectivity (concatenation operation) of the input feature maps from all the preceding layers.  
Similar also to the 2D DenseNet, the connection between every two dense blocks in ST-DenseNet is done via transition layers (TL). TL is comprised of two internal consecutive layers: 3D convolution and 3D pooling layers to resize the feature maps between dense blocks. Given the deep structure of ST-DenseNet and the added advantage of the dense connectivity, the number of input feature maps for each layer increases largely. In the conventional 2D DenseNets, two approaches were introduced for overcoming this problem. The first one is the growth rate parameter, which is used inside each dense block to control the number of feature maps generated from each layer in dense blocks. We extend the same growth rate parameter in our ST-DenseNet and we used a value of 24 for it. The second approach was the bottleneck layers. Bottleneck layers were firstly introduced in~\cite{szegedy2016rethinking} for addressing the same issue of high number of feature maps in the inception and ResNet architectures. The bottleneck layer in the case of 2D ConvNet models is a ``1 $\times$ 1 conv'' layer to reduce the number of input feature maps. In our ST-DenseNet model we will extend it to be a 3D bottleneck layer ``1 $\times$ 1 $\times$ 1 conv'' and will be at the start of each dense block. The total number of dense blocks we used in our ST-DenseNet model for the intended action prediction task are 3 blocks with 4 layers within each one. The input to the model is a sequence of the cropped BBox images (resized to 100H$\times$ 100W) of the pedestrians tracked over the past 16 frames ($\approx$ 0.5 sec of 30 FPS camera) from the multi-pedestrian tracking stage. The output is two softmax classification probability scores (to cross or not) for each input sequence of the unique tracked pedestrians. The full details of our ST-DenseNet model and each layer's spatial and temporal size are presented in Table~\ref{st-DenseNet}.
\begin{table}[t]
	\centering
	\resizebox{0.85\columnwidth}{!}{%
		\begin{tabular}{c|c|c}
			\hline
			Layers                                                                          & Output Size     & Proposed ST-DenseNet                                                                           \\ \hline
			Convolution-3D                                                                     & \cross{50} $\times$ 16 & \cross{7} $\times$ 7 conv, stride 2                                                                                                                                                                                                                                                                                                                                                                                                    \\ \hline
			Pooling-3D                                                                         & \cross{25}  $\times$ 16 & 
			\cross{3} $\times$ 3 average pool, stride 2                                                                                                                                                                                                                                                                                                                                                                                               \\ \hline
			\begin{tabular}[c]{@{}c@{}}Dense-Block-3D\\ (1)\end{tabular}                       & \cross{25}  $\times$ 16 & \multicolumn{1}{c}{\conv{4}} \\ \hline
			\multirow{2}{*}{\begin{tabular}[c]{@{}c@{}}Transition-Layer-3D\\ (1)\end{tabular}} & \cross{25} $\times$ 16  & \cross{1} $\times$ 1 conv                                                                                                                                                                                                                                                                                                                                                                                                              \\ \cline{2-3}
			& \cross{13} $\times$ 8  & \cross{2} $\times$ 2 average pool, stride 2                                                                                                                                                                                                                                                                                                                                                                                             \\ \hline
			\begin{tabular}[c]{@{}c@{}}Dense-Block-3D\\ (2)\end{tabular}                       & \cross{13} $\times$ 8  & \multicolumn{1}{c}{\conv{4}}  \\ \hline
			\multirow{2}{*}{\begin{tabular}[c]{@{}c@{}}Transition-Layer-3D\\ (2)\end{tabular}} & \cross{13} $\times$ 8  & \cross{1} $\times$ 1 conv                                                                                                                                                                                                                                                                                                                                                                                                              \\ \cline{2-3}
			& \cross{7} $\times$ 4   & \cross{2} $\times$ 2 average pool, stride 2                                                                                                                                                                                                                                                                                                                                                                                             \\ \hline
			\begin{tabular}[c]{@{}c@{}}Dense-Block-3D\\ (3)\end{tabular}                       & \cross{7} $\times$ 4   & \multicolumn{1}{c}{\conv{4}} \\ \hline
			\multirow{2}{*}{\begin{tabular}[c]{@{}c@{}}Classification Layer\end{tabular}} & \cross{1} $\times$ 1 & \cross{7} $\times$ 4 average pool                                                                                                                                                                                                                                                                                                                                                                                            \\ \cline{2-3}
			&                 & 2D fully-connected, softmax                                                                                                                                                                                                                                                                                                                                                                                              \\ \hline
		\end{tabular}
	}
	\vspace{1 ex}
	\caption{ST-DenseNet architectures for pedestrian intent action prediction task. The growth rate for all the networks is $k=24$. Note that each ``conv'' layer shown in the table corresponds the sequence BN-ReLU-3DConv. }
	\label{st-DenseNet}
	\vspace{-3 ex}
\end{table}

\section{Experiments and Results}\label{expr}
In this section, we will discuss the experiments we have done and datasets we have used for evaluating the performance of our proposed framework for real-time intent prediction of pedestrians from only video sequences.
\subsection{Training and Testing Dataset}
For training our proposed framework for the pedestrian intent prediction task, we had two separate training phases. The first one is for the ST-DenseNet model. The other training phase was for the YOLOV3 object detection model of our multi-pedestrian tracking stage. Before we get into the details of the data we used for training and evaluating the two phases, it is worth noting that during the inference/testing phase, the two stages of our framework work consecutively on the same GPU.

\subsubsection*{\textbf{Training ST-DenseNet}}
In our ST-DenseNet model, we trained it using the recently released Joint Attention for Autonomous Driving (JAAD) dataset~\cite{rasouli2017they}. JAAD dataset is vehicle-based dataset captured using a dash camera (at 30 FPS with 1080H$\times$1920W). JAAD is not only annotated with pedestrians' BBoxes but also with temporal behavioral annotations as well. The temporal behavioral annotations are such as the crossing action between a specific start and end frames. Additionally, JAAD was collected in naturalistic driving sessions across different countries (mainly in North America and Eastern Europe) under various weather conditions. The dataset consists of a total of 346 video sequences, the duration of each is 5-10 sec involving pedestrians in urban traffic environments (sample frames of the dataset are shown in Fig.~\ref{fig:sample}). The density of the annotated pedestrians in the video is relatively large with 2793 unique pedestrians with only 868 provided with behavior annotations. We have split the dataset videos into 70\% for training, 15\% for validation and 15\% for testing. The training of our ST-DenseNet was done using the training split videos with the ground truth behavior and BBoxes annotations. The labels for the pedestrians' intended action of interest (crossing or not) are done according to the formulation we described in Section~\ref{formulation}. Regarding the crossing intended action label, we crop and resize all the BBoxes sequences to (100H$\times$100W) of unique pedestrians in all frames preceding the starting frame where the pedestrian commence the crossing action. Then, we label this whole sequence of frames as the intended crossing action. For the labels of non-crossing intended actions, we chose the instances of pedestrians standing or walking beside the curb and we both cropped and resized the corresponding pedestrians' BBoxes sequence to (100H$\times$100W) for each video. Given that we are interested in predicting the intended action of pedestrians at least half a second before the committed action. Thus, we have to further pre-process the extracted sequences by running a sliding window of length 16 (roughly 0.5 sec) over all the extracted sequences before we input them to our ST-DenseNet model. As a result of this pre-processing stages, we ended up with a total number of 3602 sequence samples (16 BBoxes each) with 3061 for training and validation; 541 for testing. We trained our ST-DenseNet model using the Adam optimizer with a learning rate of 0.01 and batch size of 10 samples for 70 training epochs on Nvidia Titan X GPU. 
\subsubsection*{\textbf{Training YOLOv3}}
For training the YOLOv3, we did not start training the model from scratch instead we took advantage of the transfer learning feature of deep ConvNet models to reduce the amount of data and time needed for training as well as enhancing the model capabilities~\cite{yosinski2014transferable}. We have fine-tuned the original YOLOv3 model which was trained on MS COCO dataset~\cite{lin2014microsoft} using the same training/validation split of the JAAD dataset similar to our ST-DenseNet model. We extracted the raw images of the training split videos and resized them to (416H $\times$ 416W) to comply with the original YOLOv3 model. We ended up with a roughly 10K images annotated with at least one pedestrian BBox for each image.
\subsection{Performance of Intent Action Prediction Framework}
\begin{figure}[t]
		\includegraphics[keepaspectratio=true, width=\columnwidth]{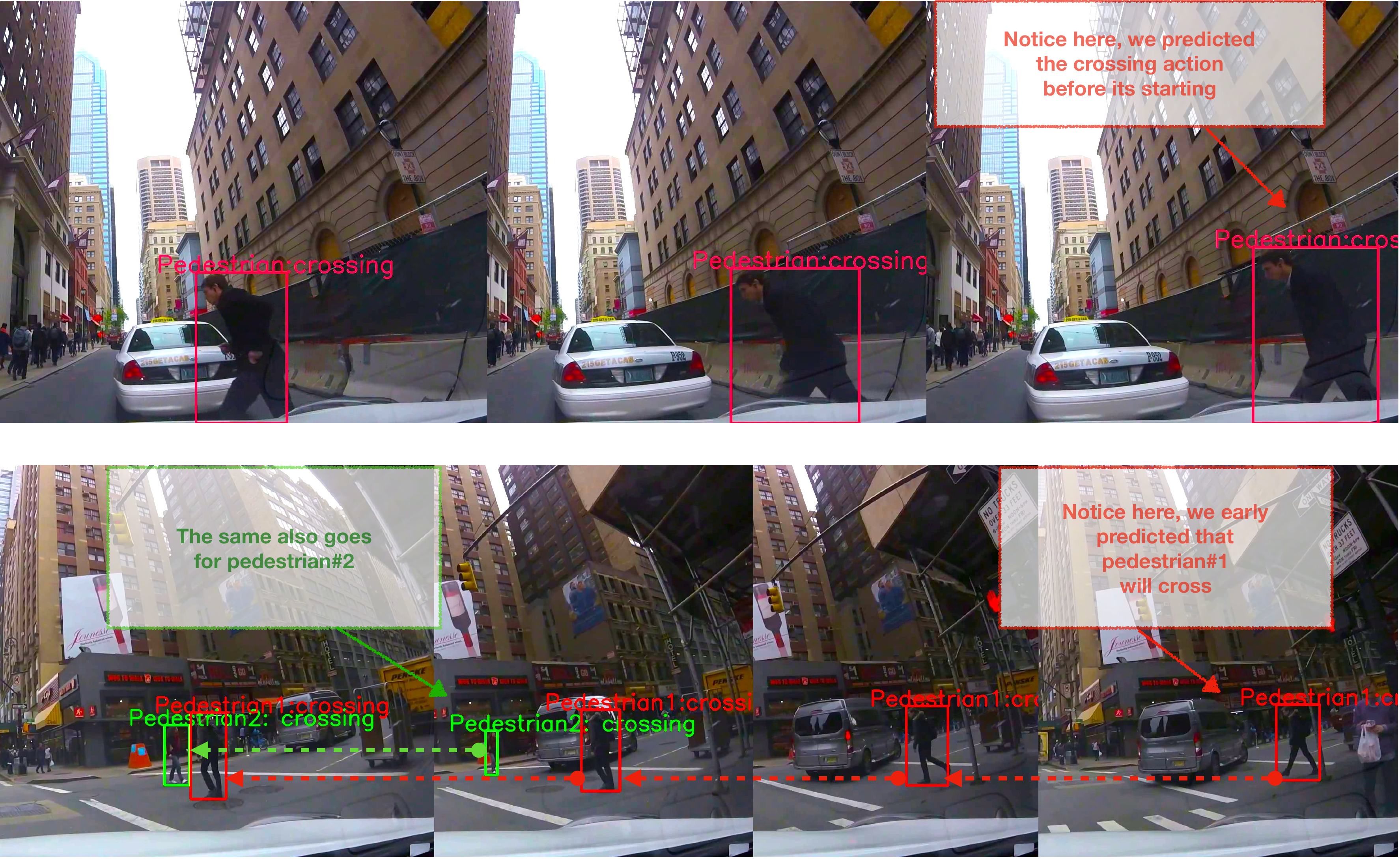}
		\caption{Qualitative results of the predictions of our framework on two scenario from the JAAD dataset. Notice how our framework can accurately predict the intended actions of the pedestrians prior to the actual action.}
		\label{fig:quali}
	\end{figure}

In this experiment, the performance of our proposed framework will be assessed in accordance with our first claim that our ST-DenseNet model can effectively predict the intended actions of pedestrians. In Table~\ref{tab:table1}, we are evaluating the performance of our framework in comparison to four baseline models; one of them was the best performing baseline model on the JAAD dataset~\cite{rasouli2017they}. In order to have a fair comparison with~\cite{rasouli2017they}, in this experiment, we are testing the four models on the same JAAD testing split using the ground truth detection and tracking without using our multi-pedestrian tracking stage. 
	\begin{table}[t]
		\caption{Comparison between out ST-DenseNet model and other baseline approaches from the literature to evaluate the performance of our proposed framework.}
		\label{tab:table1}
		\resizebox{0.85\columnwidth}{!}{%
		\centering
		\begin{tabular}{l|c|c|c|c}
			\toprule
			Approach & \multicolumn{4}{c}{Average Precision (\%)} \\
			\midrule
			& GT & ACF & SSD & YOLOv3 (ours)\\	
			\midrule
			ConvNet-Softmax~\cite{saleh2017early}        &  78.38 &  59.52 & 54.98 & 66.27\\
			ConvNet-SVM~\cite{rasouli2017they}    &  75.63 & 57.68 & 55.42 & 64.25 \\
			ConvNet-LSTM~\cite{carreira2017quo}     &  81.01 & \textbf{63.84} & 61.66 & 68.54 \\
			C3D~\cite{carreira2017quo}    &  76.83 &  51.72 & 51.66 & 56.81\\
			ST-DenseNet (ours)           &  \textbf{84.76} & 62.35 & \textbf{62.53} & \textbf{73.78}\\
			\bottomrule
		\end{tabular}
		}
	\end{table}
We have used the average precision (AP) score as our evaluation metric which is commonly utilized and accepted in the spatio-temporal action recognition and localization tasks~\cite{rasouli2017they,saleh2017early}. The AP score is a summarization of the precision-recall curve in terms of a weighted average of precisions at different threshold values between 0.0 and 1.0. The first baseline approach is the approach introduced in~\cite{saleh2017early}. The model is based on a ConvNet architecture similar to the inception architecture. For a fair comparison, we have made a slight modification to the original model to accept a temporal sequence of BBoxes rather than one single BBox. That model in return extracts the features of the sequence BBoxes after the last fully connected layer. Using these features, they are finally passed to a 2D softmax classifier. We refer to this approach as the ConvNet-Softmax. The second baseline approach is an implementation of the model introduced recently and tested on the JAAD dataset in~\cite{rasouli2017they}. The model is based on AlexNet ConvNet architecture, which takes a sequence of length 16 as pedestrian BBoxes and extracts their features to train an SVM model. This model was trained on the same training split as our ST-DenseNet model. We refer to this model as ConvNet-SVM. The third baseline approach is similar to the first baseline model but instead of feeding the extracted features to a softmax directly, they are used to train a two-layers LSTM model. The last baseline approach is the C3D approach described in Section~\ref{related}. It consists of total 9 3D convolution layers interleaved with 5 3D max pooling layers with two fully connected layers on top of them before feeding lastly to a softmax layer. Similar to our ST-DenseNet model, all baseline models were trained using an input sequence of 16 BBoxes of tracked pedestrians in the scene using the same training/validation splits. In Table~\ref{tab:table1}-second column (GT), we report the AP scores in \% for each baseline approach in comparison to our ST-DenseNet. These scores are the result of evaluating all the models on ground-truth detections/tracks (hence the name) over the same testing split of the JAAD dataset. As it can be noticed from the table our ST-DenseNet model has achieved a high AP score of 84.76\%. The nearest model with AP score to our ST-DenseNet model is the ConvNet-LSTM model, which makes a lot of sense given the recent successes of LSTM to be good in capturing the dependency in spatio-temporal data. Similar also to~\cite{carreira2017quo}, the C3D model scored lower than the ConvNet-LSTM in the AP score. Additionally, in Fig.~\ref{fig:quali}, we show sample of the predictions of our framework on two different scenarios from the JAAD dataset.   

\subsection{Effect of Noisy Observations}
In order to further evaluate the performance of our proposed framework and to support our claim regarding the second contribution of this paper, we design the following experiment. Since AGVs rely on sensor observations to get the detection and tracks of objects in the scene which are usually noisy whether due to the optics of the sensor itself or the uncertainty in the detection and tracking stages. Thus, we have trained two different object detection models other than our proposed YOLOv3 to assess the resiliency of our ST-DenseNet model against noisy observations. We used the same SORT-UKF algorithm discussed in Section~\ref{multi} of our multi-pedestrian tracking stage for all compared baseline approaches. The first object detection model is the single shot multi-box detector (SSD), is another single stage ConvNet-based model similar to YOLOv3. We specifically used the SSD-MobileNet architecture which provides a decent detections with real-time performance as it was targeted for mobile devices~\cite{huang2017speed}. The other object detection model, is the  aggregate channel features (ACF) model, another famous and heavily used model for the task of pedestrian detection in the literature and it has relatively near real-time performance on CPU. For the ACF we did not train it on JAAD training dataset due to memory and computations restrictions since it can only be trained on CPU. We however, used the ACF model trained on the Caltech pedestrian dataset from~\cite{Dollar2012PAMI}. As it is shown in Table~\ref{tab:table1}, our ST-DenseNet model continued to provide resilient results in comparison to the baseline models despite of the noisy observations. The only exception was with the ACF detector, where the ConvNet-LSTM model achieved a marginal improvement of only 1\% in AP score over our ST-DenseNet model. Overall, the best AP score achieved was with our proposed framework (ST-DenseNet+YOLOv3) with 73.78\% AP score. 

\begin{table}[t]
	\caption{Runtime performance analysis of our proposed framework in comparison to other baseline approaches.}
	\label{tab:rt}
	\resizebox{0.85\columnwidth}{!}{%
	\centering
	\begin{tabular}{c|c|c|c|c}
		\toprule
		Approach & Tracking & Prediction & Total & FPS\\
		\toprule  
		ConvNet-Softmax~\cite{saleh2017early} & 40ms & 28ms & 68ms & 14.7 \\
		ConvNet-SVM~\cite{rasouli2017they}  & 40ms & 27ms & 67ms & 14.9 \\
		ConvNet-LSTM~\cite{carreira2017quo} & 40ms & 40ms & 80ms & 12.5 \\
		C3D~\cite{carreira2017quo} & 40ms & 27ms & 67ms & 14.9\\
		ST-DenseNet (ours) & 40ms & \textbf{10ms} &  \textbf{50ms} &  \textbf{20}\\
		\bottomrule
	\end{tabular}
	}
\end{table}	
\subsection{Runtime Analysis}
In Table~\ref{tab:rt}, the run time performance of our proposed framework in comparison to the other baseline approaches is listed. Since the baseline models from the literature did not include the detection or the tracking stages as part of their action prediction models, we report the runtime performance in ms for each separate stage. The tracking column represents the first stage of the framework which is the multi-pedestrian tracking using YOLOv3+SORT-UKF. The prediction column represents the second stage which could be any of the four baseline models in addition to our ST-DenseNet model from the first column. As it can be noticed, our framework has achieved a real-time performance of 20 FPS, while the other baseline models were facing some challenges with the nearest model achieving only 14.9 FPS. All the run-time analysis experiments run on the same PC with an Intel i7 CPU and an Nvidia Titan X GPU.

\section{Conclusion }\label{concl}
In this work, we have introduced a real-time framework for the task of intent action prediction of pedestrians in urban traffic environments. We have extended state-of-the-art deep DenseNet architecture to accommodate spatio-temporal image sequences from a monocular RGB camera to predict pedestrians' intended actions. Our framework has achieved a remarkable results in comparison to a number of baseline approaches from the literature. We have scored an average precision score of 84.76\% in comparison to 76.83\% reported by the C3D model from the literature. Additionally, our framework has also achieved a real-time performance of 20 FPS.

	\bibliographystyle{IEEEtran}
	\bibliography{IEEEabrv,Ref}
	
\end{document}